\pdfoutput=1
\documentclass[11pt]{article}

\usepackage{authblk}
\usepackage{acl}

\usepackage{times}
\usepackage{float}
\usepackage{latexsym}
\usepackage{multirow}
\usepackage{graphicx}
\usepackage{url}
\usepackage{hyperref}
\usepackage{amsmath}
\usepackage{caption}
\usepackage[utf8]{inputenc}
\usepackage{stfloats}

\usepackage[T1]{fontenc}
\usepackage[utf8]{inputenc}

\usepackage{microtype}

\title{GPT-D: Inducing Dementia-related Linguistic Anomalies by Deliberate Degradation of Artificial Neural Language Models}

\author[1]{\textbf{Changye Li}}
\author[2]{\textbf{David Knopman}}
\author[3]{\textbf{Weizhe Xu}}
\author[3]{\textbf{Trevor Cohen}}
\author[4]{\textbf{Serguei Pakhomov}}
\affil[1]{Institute of Health Informatics, University of Minnesota}
\affil[2]{Mayo Clinic, Rochester, MN}
\affil[3]{Biomedical and Health Informatics, University of Washington}
\affil[4]{Pharmaceutical Care and Health Systems, University of Minnesota}

\affil[1, 4]{\textit {\{lixx3013, pakh0002\}@umn.edu}}
\affil[2]{\textit {\{knopman\}@mayo.edu}}
\affil[3]{\textit {\{xuweizhe, cohenta\}@uw.edu}}

\begin{document}
\maketitle
\begin{abstract}
Deep learning (DL) techniques involving fine-tuning large numbers of model parameters have delivered impressive performance on the task of discriminating between language produced by cognitively healthy individuals, and those with Alzheimer's disease (AD). However, questions remain about their ability to generalize beyond the small reference sets that are publicly available for research. As an alternative to fitting model parameters directly, we propose a novel method by which a Transformer DL model (GPT-2) pre-trained on general English text is paired with an artificially degraded version of itself (GPT-D), to compute the ratio between these two models' \textit{perplexities} on language from cognitively healthy and impaired individuals. This technique approaches state-of-the-art performance on text data from a widely used "Cookie Theft" picture description task, and unlike established alternatives also generalizes well to spontaneous conversations. Furthermore, GPT-D generates text with characteristics known to be associated with AD, demonstrating the induction of dementia-related linguistic anomalies. Our study is a step toward better understanding of the relationships between the inner workings of generative neural language models, the language that they produce, and the deleterious effects of dementia on human speech and language characteristics.    
\end{abstract}

\section{Introduction}

Alzheimer's disease (AD) dementia affects every aspect of cognition, including language use. Over 50 million people are currently diagnosed with AD dementia, and this number is expected to triple by 2050 \citep{world2017global, patterson2018world, prince2016world}. Furthermore, over half of the individuals living with dementia are undiagnosed \citep{Lange2017}. While AD has no known cure, timely diagnosis can prevent or alleviate adverse outcomes ranging from anxiety over unexplained symptoms to family discord and catastrophic events \citep{stokes2015, boise1999, bond2005}. However, diagnosis of AD dementia is time-consuming and challenging for patients and physicians alike, and currently relies on patient and caregiver reports, extensive neuropsychological examinations, and invasive imaging and diagnostic procedures \citep{patterson2018world}. %Individuals with unrecognized dementia suffer adverse outcomes ranging from anxiety over unexplained symptoms to family discord and catastrophic events \citep{stokes2015, boise1999, bond2005}. Findings from a large (n=2,678) international survey of attitudes towards AD dementia diagnosis show that the vast majority (>80\%) would prefer to know if their unexplained symptoms of confusion or memory loss were due to AD dementia \citep{blendon2011}. With an estimated 30-40\% of healthy adults subjectively reporting regular forgetfulness  \citep{Ponds1997,Bebbington2011} and the rapidly increasing cost of dementia treatment (estimated to exceed \$2 trillion US dollars by 2030 \citep{patterson2018world}), Thus, accurate, easy-to-use, safe and cost-effective tools for monitoring AD-related cognitive markers are urgently needed for those individuals seeking assessment of their cognitive function, or undergoing clinical evaluation for other reasons. 
Automated analysis of spoken language can potentially provide accurate, easy-to-use, safe and cost-effective tools for monitoring AD-related cognitive markers. In particular, studies have demonstrated that supervised machine learning methods can learn to differentiate accurately between patients with dementia and healthy controls \citep{fraser2016linguistic,orimaye2017predicting}, with particularly strong performance from recent deep learning (DL) models \citep{balagopalan2020bert,Roshanzamir2021TransformerbasedDN}. However, the large number of parameters employed in DL presents a danger of overfitting to the small datasets concerned, and hinders interpretability of model predictions - both critical concerns for clinical artificial intelligence applications \citep{graham2020artificial}.

As an alternative to fitting model parameters directly, we propose a novel method by which a pre-trained Transformer \citep{NIPS2017_3f5ee243} model, GPT-2 \citep{radford2019language} is paired with an artificially degraded version of itself (GPT-D), to compute the ratio of model perplexities on language from cognitively healthy and impaired individuals. We anticipate that semantic information lost with dementia progression may be localized to particular layers of a neural language model, and that one can simulate this information loss by systematically modifying parameters in these layers. Specifically, we hypothesize that impairing certain layers of a DL model can result in linguistic deficits that are also observed in dementia. We further hypothesize that unlike prior work fitting model parameters to labeled ``Cooke Theft'' transcripts, this approach will detect task-agnostic linguistic anomalies, permitting evaluation of language from casual conversations. We evaluate these hypotheses by targeting individual layers for induction of dementia-related linguistic anomalies, resulting in a degraded model -- GPT-D. We then assess the ability of a paired perplexity approach combining GPT-2 with GPT-D to identify transcripts from participants with dementia. In addition, we assess generalization performance, and consider the extent to which the best-performing degraded model reflects linguistic anomalies known to occur in AD dementia: usage of higher frequency words, and repetitiveness. The contributions of this work can be summarized as follows: a) we develop a novel method for automated detection of dementia-related linguistic anomalies, involving deliberate degradation of a pre-trained Transformer model; b) this method exhibits state-of-the-art (SOTA) within-set performance for models trained on text alone, and is distinguished by its ability to generalize from cognitive tasks to conversational data; c) the degradation process induces linguistic anomalies observed in dementia in language generated by GPT-D\footnote{Our code is available at \url{https://github.com/LinguisticAnomalies/hammer-nets}}.

%Our results suggest that the paired perplexity approach using the ratio between the GPT-2 and GPT-D model perplexities approaches SOTA performance \citep{cohen-pakhomov-2020-tale} for discriminating between language samples of patients with dementia and controls across different tasks and discourses \textit{without} introducing more data. Furthermore, our methods generalize beyond the reference sets they are trained on, greatly expanding their scope of application. We demonstrate this is not the case with previously investigated methods based on fine-tuning a pre-trained Bidirectional Encoder Representations from Transformer (BERT) model \citep{devlin-etal-2019-bert}. We also observe significant differences in lexical frequency and type-to-token ratio characteristics of text generated by GPT-2 and GPT-D, which aligns with previous studies showing that language of dementia patients is sensitive to lexical frequency effects \citep{cohen-pakhomov-2020-tale,pekkala2013}. 

\section{Background}

Building on a rich body of evidence that machine learning methods can learn to distinguish between language from healthy controls and dementia patients (for a review, see \citet{lyu2018review, petti2020systematic}), recent work leveraging pre-trained Transformer models has demonstrated improvements in performance over prior approaches. %\citet{Roshanzamir2021TransformerbasedDN} compared multiple Transformer models, including BERT and XLNet \citep{NEURIPS2019_dc6a7e65} on the Dementia Bank dataset (Pitt corpus) (DB) \citep{becker1994natural}, and report an accuracy of 88.08\%. This   
\citet{balagopalan2020bert} fine-tuned the BERT  \citep{devlin-etal-2019-bert} model on the training set of the AD Recognition through Spontaneous Speech (ADReSS) Challenge \citep{bib:LuzHaiderEtAl20ADReSS}, which was developed, in part, to address the lack of standardized train/test splits and subset definitions in prior work using DementiaBank \citep{becker1994natural} (DB). \citet{balagopalan2020bert} report an accuracy of 83.3\% on the test set, an improvement over  machine learning models with expert-defined features. Performance can also be further boosted by introducing more data from the same picture description task \citep{guo2021crossing}. These findings suggest a promising direction, as models can be developed without extensive feature engineering. However, additional task-specific data are not always available. DL models with millions of parameters are vulnerable to overfitting with small data sets, which may be difficult to detect as they are hard to interpret. %found that BERT performance on the ADReSS set benefits from training  on additional transcripts with similar content, specifically 1,366 cookie theft transcripts with ``noisy'' labels derived from verbal fluency scores with education-adjusted thresholds, drawn from the Wisconsin Longitudinal Study (WLS) \citep{herd2014cohort}. %model with \texttt{BCEWithLogitsLoss}, where more than one label can be assigned (i.e., diagnosis=[case\textbar control], source = [WLS\textbar ADReSS]), 

%Models with millions or billions of parameters are hard to interpret. However, in some cases 
However, some DL models can be distilled into a single interpretable feature: language model (LM) perplexity (PPL). PPL is a measurement of how well a language sample fits a trained LM. Intuitively, a model trained on language from cognitively healthy participants should be ``surprised'' by language from participants with dementia, and the opposite should also be true. Accordingly, the difference between the \textit{paired perplexities} from ``cognitively healthy'' and ``dementia'' language models produces SOTA results on the task of identifying transcripts from participants with dementia \citep{fritsch2019automatic, cohen-pakhomov-2020-tale}, effectively condensing neural network parameters to a single diagnostically useful feature. Contemporary deep LMs such as GPT-2 are \textit{already} trained on large amounts of text, that has presumably been authored predominantly by cognitively healthy individuals. The difficulty with leveraging these models within the paired perplexity paradigm arises from the lack of a correspondingly large set of text from participants with dementia. We negotiate this difficulty by deliberately degrading a Transformer model to limit its semantic processing capabilities, obviating the need for large amounts of dementia-specific training data. We show that the resulting models can effectively identify transcripts from participants with dementia, generalize across language samples and tasks, and generate text with linguistic characteristics of this condition. 

\section{Methods}

\begin{table*}[ht]
\small
\begin{tabular}{|l|l|l|l|l|l|l|l|}
\hline
\multicolumn{2}{|l|}{\textbf{Dataset}} & \multicolumn{3}{l|}{\textbf{Dementia}} & \multicolumn{3}{l|}{\textbf{Healthy Controls}} \\ \hline
\multicolumn{2}{|l|}{} &   \multicolumn{1}{|p{1.6cm}|}{\centering N \\ participants}    &  \multicolumn{1}{|p{1.8cm}|}{\centering MMSE \\ Mean (SD) }    &  \multicolumn{1}{|p{1.8cm}|}{\centering Transcript \\ length \\ Mean (SD)}     & \multicolumn{1}{|p{1.6cm}|}{\centering N \\ participants}     &  \multicolumn{1}{|p{1.8cm}|}{\centering MMSE \\ Mean (SD) }    &  \multicolumn{1}{|p{1.8cm}|}{\centering Transcript \\ length \\ Mean (SD)} \\ \hline
\multirow{3}{*}{\textbf{ADReSS}}  & train  &   54    &   17.1 (5.5)    & 104 (63) & 54     &   29.1 (1.9)    &    114 (49)        \\ \cline{2-8} 
                   & test  &  24     &   19.5 (5.4)    &    95 (47)   &  24     &  28.8 (1.5)     &  120 (72)    \\ \cline{2-8} 
                    & all  &   78    &   17.8 (5.5)    &    101 (58)   &   78    &  29 (1.2)    &  116 (56)    \\ \hline
\multicolumn{2}{|l|}{\textbf{DB}} &   169    &   20.2 (4.6)    &  959 (534)     &   99    &   29.1 (1.1)    &  1085 (556)    \\ \hline
\multicolumn{2}{|l|}{\textbf{CCC}} &   234    &   NA    &  1213 (943)     &   48    &    NA   &   714 (308)  \\ \hline
\end{tabular}
\caption{Basic characteristics of datasets}
\label{tab:data}
\end{table*}

\subsection{Data}
We used three publicly available datasets\footnote{While the data used in this paper are publicly available, we are not able to redistribute any of these data as per Data Use agreement with Dementia Bank and the Carolinas Conversation Collection.}: DB, ADReSS, and the Carolinas Conversation Collection (CCC) \citep{pope2011finding}. Dataset characteristics are provided in Table~\ref{tab:data}. DB is a publicly available compendium of manually transcribed audio recordings of neuropsychological tests administered to healthy participants and patients with dementia. A detailed description is available in \citet{becker1994natural}. In brief, the tests include a picture description task from the Boston Diagnostic Aphasia Examination \citep{goodglass1983boston}, a widely-used diagnostic test for language abnormality detection. In this task, the participants are presented with a ``Cookie Theft'' picture stimulus (see Figure~\ref{fig:cookie_theft} in Appendix), %(a scene in which two children collude to steal cookies from a cabinet while their mother is distracted), 
and are asked to describe everything they see occurring in the picture. In other words, DB data are from tasks that were explicitly designed to detect language abnormalities in dementia patients. We restricted the original set of 194 participants with any AD diagnosis only to those that were assessed as having probable AD, resulting in a set of 169 patients and 99 controls. The ADReSS set is a subset of DB, which the controls and dementia participants were matched age and gender, resulting in a balanced dataset consisting of a total of 156 samples (78 with dementia and 78 controls) split into training and testing portions. Unlike the two preceding datasets derived from picture description tasks, CCC is a collection of 646 transcribed recordings of interviews of 48 elderly cognitively normal individuals with non-dementia related chronic conditions, and 234 individuals with a diagnosis of dementia. Interview topics vary considerably, and include discussions of the participant's health.

Additionally, we used a set of six synthetic ``Cookie Theft'' picture description narratives created by \citeauthor{bird2000rise} (\citeyear{bird2000rise}) to study the impact of semantic dementia on verb and noun use in picture description tasks. %While \citeauthor{bird2000rise} focused on semantic dementia, a distinct condition from AD dementia, these synthetic narratives were not based on patients with semantic dementia. Rather, 
The transcripts were created to manipulate lexical frequency (which is also relevant in AD dementia, where words with higher lexical frequency tend to feature prominently \citep{almor1999alzheimer}) by first compiling a composite baseline narrative from samples by healthy subjects, and then removing and/or replacing nouns and verbs in that baseline with words of higher lexical frequency (e.g., ``mother'' vs. ``woman'' vs. ``she''). Lexical frequency was calculated using the Celex Lexical Database (LDC96L14) and words were aggregated into groups based on four log frequency bands (0.5 - 1.0, 1.0 - 1.5, 1.5 - 2.0, 2.5 - 3.0: e.g., words in the 0.5 - 1.0 band  occur in Celex more than 10 times per million). We used these synthetic data to help with interpretation of the effects resulting from artificially impairing the GPT-2 model.

We performed basic pre-processing of transcripts in each dataset by which we removed speech artifact descriptions %(e.g. ``background noise'', ``chuckles'', ``clears throat'') 
and converted non-ASCII characters to plain text. We also excluded portions of transcripts that represented speech that did not belong to the participant. %(e.g., interviewer or test administrator speech).

% CL: do we need to cite Celex here?

%\subsection{Text Pre-processing}

\subsection{Modeling and Evaluation}
We evaluated models for \textit{classification} performance using the standard ADDReSS train/test splits. We then performed \textit{cross-validation} of GPT-D models to assess the stability of the best-performing configurations across folds. For \textit{generalization} performance, we evaluated how well models trained on one corpus performed on others. We also assessed differences in text \textit{generation} between GPT-2 and GPT-D, by estimating repetitiveness and lexical frequency, as well as through salience-based \textit{visualization}.

\subsubsection{Artificial Impairment: Locations}

\begin{figure*}[ht]
    \centering
    \includegraphics[width=0.7\textwidth]{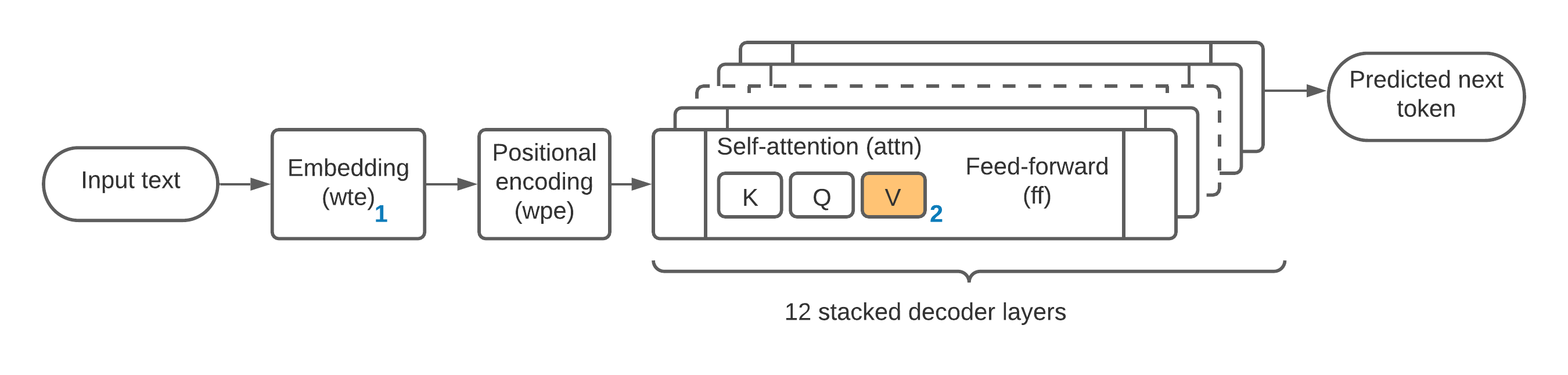}
    \caption{Impairment locations within the GPT-2 (small) model.}
    \label{fig:GPT-2-impairment-locations}
\end{figure*}

We experimented with impairing the GPT-2 (small) model in two locations as illustrated in Figure~\ref{fig:GPT-2-impairment-locations} with various portions. We found that impairing 50\% of values in the corresponding location resulting in generally better performance, among 25\%, 50\%, 75\% and 100\% impairment. The \emph{embedding layer} (see (1) in Figure \ref{fig:GPT-2-impairment-locations}) is a 50,257$\times$768 matrix where each row represents a token in the model's vocabulary. The embedding layer was impaired by randomly masking 50\% of the rows of of the embedding matrix.
%was impaired by randomly modifying a portion of the rows in the embedding matrix in which each row is a distributed representation of the meaning of a token in the model's vocabulary. This was done by replacing a randomly selected 50\% of the 50,257 components of each of the rows with vectors containing zeroes. 
The \emph{self-attention mechanism} (denoted (2) in Figure \ref{fig:GPT-2-impairment-locations}) was impaired by masking the first 50\% of columns in the Value matrix of the concatenated Query-Key-Value matrices. We found that masking random columns resulted in worse performance in preliminary experiments.

The self-attention mechanism multiplies vectors representing an input sequence by three identically-sized matrices, namely Query (Q), Key (K) and Value (V) each with dimension ($d$) of 768$\times$768. Q generates a representation of the current token which is compared with token representations derived from K, to calculate each token's influence on the contextual representation of the current one. Multiplying by V generates a semantic representation of each token, which is added to the outgoing representation of the current token in accordance with this influence. The attention weights are calculated by Equation~\ref{eq:attn}, and the parameters of the matrices are updated during the training process.
\begin{equation}
    attention(Q,K,V)= softmax(\frac{QK^{T}}{\sqrt{d}}V)
    \label{eq:attn}
\end{equation}

\noindent The GPT-2 model's attention mechanism in each of the 12 decoder layers contains 12 attention heads that are represented as vectors of 64 parameters. We impaired 50\% of those parameters of V in various combinations of attention heads in each decoder layer by masking them as zeroes. We only did this in V matrices, as their parameters directly determine the content of the vectors that are passed on to the subsequent feed-forward layer, while the Q and K matrices determine how this content is weighted when generating the representations to be propagated as weighted sums of vectors that have been transformed by the Value matrix. 

\subsubsection{Artificial Impairment: Patterns}
We also experimented with three ways of introducing artificial impairment into the attention mechanism in single and multiple decoder layers: individual, cumulative, and combination. The \emph{individual} approach was to simply impair all 12 layers one at a time. The \emph{cumulative} approach consisted of impairing decoder layers sequentially starting with the bottom decoder layer (layer 0) and adding impairment to layers above it one at a time up to layer 11, resulting in total of 12 combinations of impairments. The \emph{combination} approach consisted of impairing all possible combinations of layers, one combination at a time, resulting in 4096 combinations. The degraded models were subsequently used in combination with the original GPT-2 model to calculate the difference and ratio of PPLs between these two models on each input transcript.

\subsection{Interpretation of Neural Model Behavior}
\textbf{Classification Performance: }For the paired perplexity approach, we estimated the ratio of model PPLs ($\frac{PPL_{\text{GPT-2}}}{PPL_{\text{GPT-D}}}$) for each transcript. These PPLs were averaged for participants with multiple transcripts. All validation methods commenced with calculating the area under the receiver-operator characteristic (ROC) curve (AUC). From this, accuracy (ACC) was determined at equal error rate (EER), a threshold where the false acceptance rate and false rejection rate from an ROC curves is equal. We also calculated Pearson correlation between the ratio in perplexities of the GPT-2 and GPT-D models and the MMSE scores where available (CORR).We used the original fixed single split between training and testing data provided by the creators of the dataset to compare our results to those published by others on ADReSS. 

\textbf{Cross-validation Performance: }For all datasets (including ADReSS), we performed standard cross-validation by which we split each dataset into disjoint folds and first determined which combination of GPT-D attention layers results in best performance on the training portion of each fold and then tested that combination on the test portion of the fold averaging the AUC, ACC and CORR values (if available) across the folds. We selected 5-fold cross-validation due to the relatively small size of the ADReSS, DB, and CCC datasets. %We selected five folds to keep the standard deviation in AUC, ACC and CORR values across folds to under 0.1 (10\%). 
To ensure reproducibility across runs, data folds for cross-validation were extracted using the KFold method from the scikit-learn library \citep{scikit-learn} with shuffling and a fixed random seed.

\textbf{Generalization Performance: }We tested generalizability of the paired perplexity approach by evaluating its performance across datasets. We first determined the best-performing pattern of impairment based on the highest AUC obtained on each dataset, and then applied the model impaired with that pattern to the remaining datasets. 

\textbf{Baseline Models: }We compared our model performance on transcript \textit{classification} with the previous text-only SOTA \citep{balagopalan2020bert}, which was obtained with a 12-layer BERT model fine-tuned on the ADReSS training set, and evaluated on the test set. To evaluate the \textit{generalization} performance, we followed this work's hyperparameter choices and fine-tuned BERT and DistilBERT \citep{sanh2019distilbert}\footnote{Available on Huggingface \url{https://huggingface.co/transformers/index.html}}, a distilled BERT base model that is compact and more efficient. We fine-tuned these models on the entire ADReSS, DB and CCC datasets separately, then evaluate the three resulting models on every other set.

\textbf{Language Generation: }To prompt the GPT-2 and GPT-D models to generate text we utilized \citeauthor{bird2000rise}'s synthetic ``Cookie Theft'' picture description narrative that represents a composite of narratives produced by healthy controls. Table~\ref{tab:text_gene_comb} (in Appendix) illustrates the text generated by GPT-2 and GPT-D in response to prompt sentences taken from the synthetic narrative. Both GPT-2 and GPT-D models were induced to generate at least 20 additional tokens with a beam search \citep{wiseman-rush-2016-sequence} that keeps the top $n$ hypotheses ($n=5$ in this case) at each time step and eventually returns the sequence of hypotheses that achieved the highest probability after reaching the end-of-sequence token. Beam search also works well when the length of output is not predictable, which fits the nature of the language tasks represented by the corpora we tested. However, one of the challenges of using beam search for text generation is that it tends to generate repeated words. We added a penalty for generating repetitive unigrams and implemented the top-p algorithm \citep{welleck2019neural} to keep the set of potential words as small as possible while the cumulative probability of this set is greater than the specific probability $p$ ($p=0.9$ in our case). The penalty was applied equally to GPT-2 and GPT-D to avoid potentially biasing one of these models to produce more repetitions. After the models generated five best predictions for each prompt, we chose the first non-empty pair of outputs from both the GPT-2 and GPT-D models as the final result.

\textbf{Lexical frequency and repetitiveness: }Previous work \citep{cohen-pakhomov-2020-tale} suggests that neural language models are sensitive to lexical frequency. We investigated whether GPT-D generates content of higher lexical frequency than the GPT-2 model. To compute lexical frequency, we split each generated output into tokens with the help of the NLTK\footnote{\url{https://www.nltk.org/}}. We did not stem the tokens to avoid increasing lexical frequency by artificially merging different tokens with the same stem. In addition to the stopwords provided by NLTK, we treated tokens with following part-of-speech tags a) \textit{PRP} (personal pronoun), b) \textit{PRP\$} (possessive pronoun), c) \textit{WP\$} (possessive wh-pronoun), and d) \textit{EX} (existential there) as stopwrods. We also added the \textit{n\'\;t} token and tokens starting with \textit{\'} to the list of stopwords. 
Log lexical frequency of each qualified generated token was calculated based on occurrence in the SUBTLEX$_{\text{us}}$ corpus \citep{brysbaert2009moving}. Tokens that do not appear in SUBTLEX$_{\text{us}}$, were removed as out-of-vocabulary (OOV) items. To asses the degree of repetition present in the generated text, we calculated the type-to-token ratio (TTR) as the number of word types divided by the number of word instances.

%After removing the extended stopwords from the generated text, we calculated the log lexical frequency for each token in the SUBTLEX$_{\text{us}}$ corpus \citep{brysbaert2009moving} mapped to the generated text. I

\textbf{Salience Visualization: }We used the gradient$\times$input saliency proposed in \citet{Denil2014ExtractionOS}, as implemented with the \texttt{ecco}\footnote{\url{https://github.com/jalammar/ecco}} Python package for visualization. Saliency is defined as $||\bigtriangledown_{x_{i}}f_{c}(\mathbf{x}_{1:n})\cdot x_{i}||_{2}$, which is the L2 normalized back-propagated gradient with respect to a) the dot product of the embedding vector of all previous input tokens ($\mathbf{x}_{1:n}$), and b) the model output of token $x_{i}$ ($f_{c}(\mathbf{x}_{1:n}$)), where $c$ is the predicted token at time-step $i$. A previous study \citep{serrano-smith-2019-attention} found that raw attention weights were not interpretable for any intermediate representation of a language model. Instead, \citet{bastings-filippova-2020-elephant} argued that saliency is the preferred method for interpretability as it takes the entire input into account and reveals the relevance of each input token to the next predicted token in the sequence.

\begin{figure*}[t]
    \centering
    \includegraphics[width=0.85\textwidth]{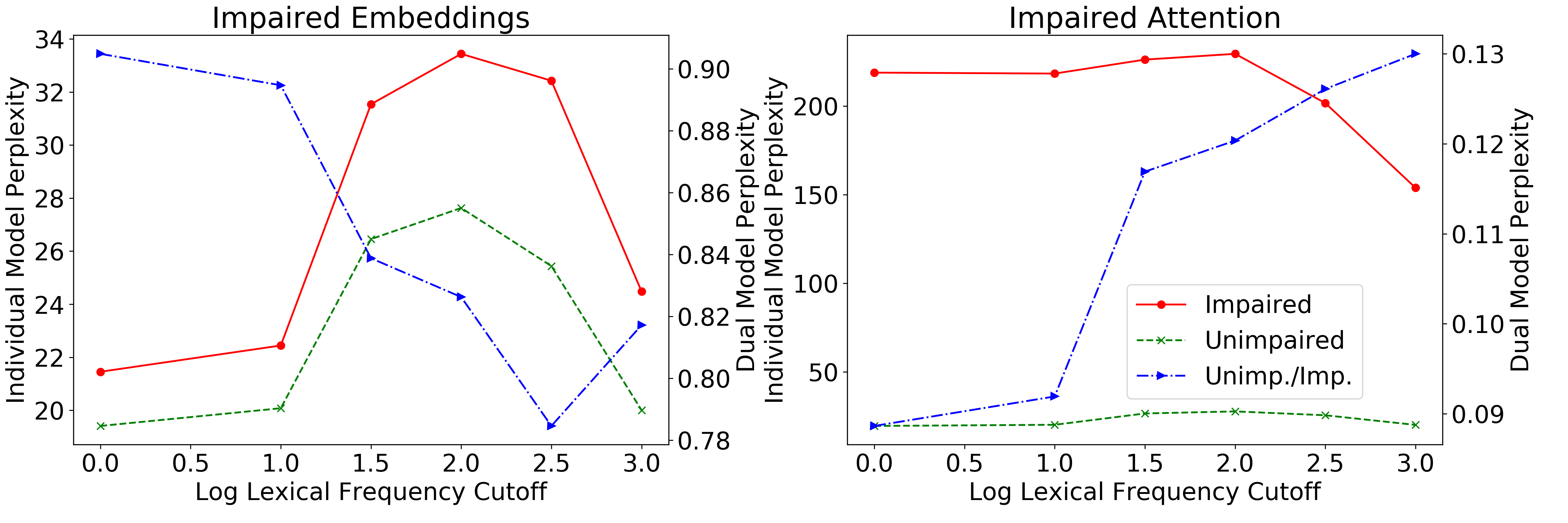}
    \caption{Effects of artificial impairment on model perplexity in synthetic picture description narratives. Higher values on the $x$ axis indicate transcripts simulating more advanced disease.}
    \label{fig:att_vs_embed}
\end{figure*}

\begin{table*}[h]
\small
\resizebox{\textwidth}{!}{
\begin{tabular}{|l|l|l|l|l|l|l|}
\hline
Dataset & \multicolumn{3}{l|}{\textbf{Combination Impairment Pattern}} & \multicolumn{3}{l|}{\textbf{Cumulative Impairment Pattern}} \\ \hline
 & AUC (SD) & ACC (SD) & r with MMSE (SD) & AUC (SD) & ACC (SD) & r with MMSE (SD) \\ \hline
ADReSS & 0.80 (0.06) & 0.71 (0.07) & -0.52 (0.08) & 0.79 (0.02) & 0.68 (0.03) & -0.51 (0.05) \\ \hline
DB & 0.81 (0.07) & 0.76 (0.04) & -0.45 (0.06) & 0.83 (0.02) & 0.73 (0.02) & -0.41 (0.14) \\ \hline
CCC & 0.77 (0.04) & 0.71 (0.04) & -- & 0.72 (0.04) & 0.64 (0.09) &--  \\ \hline
\end{tabular}}
\caption{Five-fold cross-validation results of all possible combination of impairment pattern (masking 50\% of value matrix as zeroes) for cumulative and combination methods on ADReSS, DB, and CCC.}
\label{tab:cv_result}
\end{table*}

To make the visualizations comparable for the two models, we repeatedly prompted both models with the same input until both models generated the same token as the prediction. It is worth noting that \texttt{ecco} for visualization supports limited text generation arguments compared to the \texttt{transformers} package, which we used for language generation task. Consequently, we only used the top-p algorithm currently supported by \texttt{ecco} for our visualizations.

\section{Results}

\textbf{Impairment Location: }The contrast in the effects of artificial impairment on the embedding and attention layers (locations 1 and 2 in Figure~\ref{fig:GPT-2-impairment-locations}, respectively) is illustrated in Figure~\ref{fig:att_vs_embed}. Impairing embeddings results in a distribution of perplexity values over the range of impairment in the synthetic narratives very similar to that of the GPT-2 model. Impairing attention, however, results in a sharp decrease in PPL on the more perturbed narratives (those narratives simulating more impairment), which yields a monotonically increasing step-like function over $\frac{PPL_{\text{GPT-2}}}{PPL_{\text{GPT-D}}}$ that lends itself well to thresholding for categorization. These results were confirmed by testing on available corpora the discriminative ability of the paired perplexity approach by artificially impairing only the embedding layer, which resulted in near-random AUCs (close to 0.5 - data not shown). Consequently, in subsequent results we will show attention-based models only.

\textbf{Classification Performance: }For comparison with previous work using the ADReSS dataset, the best training set performance was obtained by impairing 50\% of each attention head in layers 0-5, 6, and 8-9. This pattern achieved an AUC of 0.88 (ACC = 0.75, CORR = -0.55) on the test split. The cumulative impairment method performed slightly better. Impairing 50\% of each attention head in the first 9 layers resulted in best performance on the training set, and AUC of 0.89 (ACC = 0.85, CORR = -0.64) on the test split. We note that this accuracy exceeds the average result reported by \citet{balagopalan2020bert}, and approaches the performance of their best run. 

\textbf{Cross Validation: }The results of within-set cross-validation are summarized in Table~\ref{tab:cv_result}. Both \textit{combination} and \textit{cumulative} methods had small standard deviations ($\sim 0.1$) with over or near 0.7 mean AUC on all sets. Estimates from the paired perplexity approach for both methods were negatively correlated with MMSE on the ADReSS (-0.52, -0.51) and DB (-0.45, -0.41) sets, respectively. The best performance obtained with the \textit{individual} approach resulted in AUC of 0.66 (ACC: 0.64) with impairment of layer 8 on the DB dataset; AUC of 0.70 (ACC: 0.66) with impairment of layer 8 on the ADReSS dataset; and AUC of 0.71 (ACC: 0.63) with impairment in layer 7 on CCC.

 %We note that this accuracy exceeds the average result reported by \citep{balagopalan2020bert}, and approaches the performance of their best run. 
\begin{table}[h]
\small
\begin{tabular}{|p{2.5cm}|p{1.155cm}|p{1.155cm}|p{1.155cm}|l|}
\hline
         & \multicolumn{3}{|c|}{\textbf{Testing dataset}} \\ \hline
       \textbf{Training method}  & \textbf{ADReSS} & \textbf{DB} & \textbf{CCC} \\ \hline
              
   (Best pattern:AUC)     & AUC/ACC & AUC/ACC & AUC/ACC  \\ \hline
\multicolumn{4}{|l|}{\textbf{Cumulative Impairment Pattern}}   \\ \hline
ADReSS (0-8:0.80) & -- & -- & 0.77/0.72  \\ \hline
DB  (0-4:0.82) & -- & -- & 0.69/0.68  \\ \hline
CCC  (0-2:0.72) & 0.70/0.63 & 0.74/0.63 & --  \\ \hline
\multicolumn{4}{|l|}{\textbf{Combination Impairment Pattern}}  \\ \hline
ADReSS  ~~~~~~~~~~~~~~~~~~~(0-6,8:0.80) & -- & -- & 0.76/0.71 \\ \hline
DB  (0-6,8:0.80) & -- & -- & 0.76/0.71  \\ \hline
CCC  ~~~~~~~~~~~~~~~~~~~~~~~~~~(1-3,5,7,9-11:0.79) & 0.69/0.61 & 0.72/0.67 & --  \\ \hline
\multicolumn{4}{|l|}{\textbf{Fine-tuned BERT}}  \\ \hline
ADReSS & -- & -- & 0.64/0.63\\\hline
DB & -- & -- & 0.67/0.6\\\hline
CCC & 0.71/0.66 & 0.7/0.65 & --\\\hline
\multicolumn{4}{|l|}{\textbf{Fine-tuned DistilBERT}}  \\ \hline
ADReSS & -- & -- & 0.67/0.57\\\hline
DB & -- & -- & 0.67/0.6\\\hline
CCC & 0.65/0.62 & 0.47/0.45 & --\\\hline
\end{tabular}
%\justified
\caption{Generalizability of GPT-2/GPT-D approach compared to fine-tuning on BERT and DistilBERT. All evaluation metrics are calculated at EER rate. The best-performing pattern and its performance are indicated with parentheses and separated by colon.}
\label{tab:cross_data}
\end{table}

\begin{figure*}[!b]
    \centering
    \includegraphics[width=\textwidth]{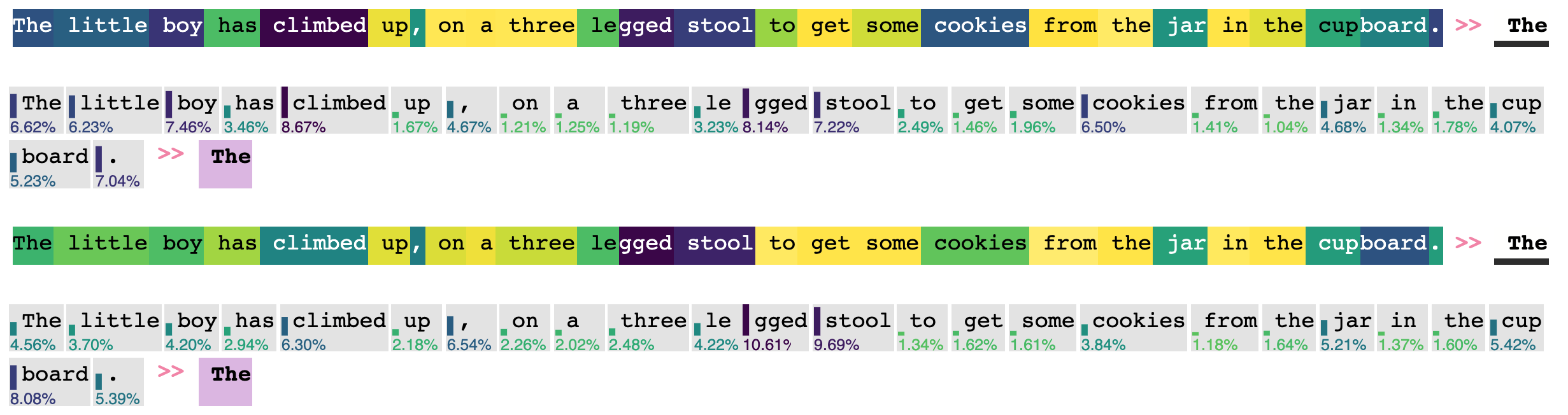}
    \caption{An informal illustration of differences in contributions of input tokens to generating the word ``The'', for GPT-2 (top) and GPT-D (bottom) models. Percentages and color represent the degree of contribution.}
    \label{fig:ecco_the_one_sent}
\end{figure*}

\textbf{Generalization: }The results of generalization evaluation are shown in Table~\ref{tab:cross_data}. Both cumulative and combination methods yielded similar performance on CCC, where both AUC and ACC were close to or exceeded 0.7. In contrast, fine-tuning BERT and DistilBERT resulted in near-random classification performance on the corresponding validation dataset. While fine-tuning BERT on conversational discourse samples in CCC and applying it to the picture descriptions in ADReSS and DB generalized well as compared to the paired perplexity approach, it did not generalize in the opposite direction when BERT was fine-tuned on ADReSS and DB picture descriptions and applied to conversations in CCC.

\textbf{Language Generation: }Table~\ref{tab:lan_stat_accumu} reports mean lexical frequency estimates for words contained in the text generated by GPT-2 and GPT-D models. The GPT-D model was induced by using the best-performing patterns of impaired layers determined from cumulative and combination methods for pattern selection on the available datasets. Both GPT-2 and GPT-D generate $\sim 1$ OOV token on average for each prompt. In general, the resulting GPT-D model generated text consisting of words with higher lexical frequency than words in the text generated by the GPT-2 model across all datasets and methods, even though some of the differences failed to reach statistical significance. All GPT-D models also generated more repetitions, evident as lower TTRs .

\begin{table}[ht]
\small
\begin{tabular}{|p{2cm}|l|l|l|l|}
\hline
Dataset \newline (Pattern)& \multicolumn{2}{l|}{\textbf{LF}} & \multicolumn{2}{l|}{\textbf{TTR}} \\ \hline
 & GPT-2 & GPT-D & GPT-2 & GPT-D \\ \hline
\multicolumn{5}{|l|}{\textbf{Cumulative}}   \\ \hline
ADReSS (0-8) & 9.48 & 9.82*  & 72\% & 50\% \\ \hline
DB (0-4) & 9.49 & 9.83*  & 72\% & 49\% \\ \hline
CCC (0-2) & 9.48 & 9.54 & 72\% & 51\% \\ \hline
\multicolumn{5}{|l|}{\textbf{Combination}}   \\ \hline
ADReSS/DB \newline (0-6,8) &  9.5 & 9.41   & 72\%  & 55\%  \\ \hline
CCC \newline (1-3,5,7,9-11) & 9.45  & 9.92**  & 73\% & 64\% \\ \hline
\end{tabular}

\caption{Mean log lexical frequency (LF) and type-to-token ratio (TTR) of the generated text. The best-performing pattern is indicated with parentheses. * indicates p-value < 0.05 and ** indicates p-value < 0.01. P-values were obtained with two-sided Welch's t-test}
\label{tab:lan_stat_accumu}
\end{table}

% cumulative
% ADR: 9 
% DB: 8
% CCC: 9
% combination
% ADR: 8
% DB: 7
% CCC: 8

\textbf{Salience Visualization: }Figure~\ref{fig:ecco_the_one_sent} shows the magnitude of the contribution for each token in the prompt used to initiate text generation to the model's prediction of the same token '\texttt{the}'. The weight of the contribution of each token is shown as a percentage that can be interpreted as the amount of contribution the model derives from it. We observe in Figure~\ref{fig:ecco_the_one_sent} that impairing GPT-2's attention heads leads to the redistribution of the model's contribution to the words in the prompt when making the prediction of the next word. For the GPT-2 model, tokens '\texttt{boy}', '\texttt{climbed}', and '\texttt{cookies}' contributed more when predicting '\texttt{the}'. However, for the GPT-D model those word tokens did not clearly stand out as substantially contributing to the prediction in either of these examples. Furthermore, tokens corresponding to function words (e.g., '\texttt{on}', '\texttt{a}' and '\texttt{from}') contributed little to the predictions generated by the GPT-2 model;  however, these tokens contributed more for predictions generated by GPT-D model. As evident in the examples in Figure~\ref{fig:ecco_the_one_sent}, the salience of the words in the prompt is much more diffuse when the GPT-D model is making the prediction - i.e. the model is uncertain with respect to what it should consider as important. In contrast, for the GPT-2 model the key elements of the ``Cookie Theft'' scenario - '\texttt{cookie}', '\texttt{three-legged stool}', '\texttt{boy}' - stand out as highly salient. These observations, although informal and qualitative, indicate that the impairment of the self-attention mechanism in GPT-2 results in a ``behavior'' resembling that observed in all stages of AD dementia as a result of impaired selective attention that in turn reduces one's ability to encode new information in episodic memory (see \citet{perry2000} for a comprehensive review).

%We further examine the hypothesis that the linguistic behavior of the GPT-D LM resembles that observed in AD dementia by examining the salience of words in the context with which the GPT-2 and GPT-D models are prompted, in relation to the first word that each of the two models generates. 

\section{Discussion}
% TODO: something worth mentioning:
% DistilBERT did not overperform BERT 
%   - smaller model does not work this specific task and not surprisingly, large model can generalize better
%   - larger model needs more resource to train, thus our approach is better and economical for blah blah blah (not sure how to write the exact implication for this)
% possible clinical application on daily conversation

Our key findings are as follows. First, we show that the paired perplexity approach using the ratio between the GPT-2 and GPT-D model perplexities approaches SOTA performance on ADReSS, leveraging GPT-2's extensive pre-training \textit{without} requiring a comparably large data set from dementia patients. Second, this approach generalizes from ``Cookie Theft'' picture description data to casual conversation, in contrast to BERT/DistilBERT fine-tuning. Finally, artificial impairment of GPT-2's self-attention induces linguistic anomalies observed in dementia. 
%The paired perplexity approach results in near-SOTA performance on DB  for discriminating between language samples of patients with dementia and controls. In prior work, fine-tuning the BERT model on the fixed training subset of the ADReSS set resulted in an average 83.3\% accuracy on the held-out test set (best performance = 85.14\%), and an average of 81.8\% accuracy in cross-validation experiments on the training set , with averages reported over three random instantiations of the model. 

The best-performing cumulative pattern for the ADReSS training set resulted in accuracy of 0.85 in the test set, exceeding the best BERT results reported on this test set ($\overline{x}$ ACC = 0.833 \citep{balagopalan2020bert}). However, our approach contrasts with approaches that train or fine-tune language models using a specific dataset, and test on held-out components of the same set. While our approach does require some labeled data through which to determine the best-performing layers to impair, our results demonstrate generalization to other datasets and populations as well as a different type of discourse - spontaneous conversations. GPT-D is reliably less perplexed by dementia-related linguistic anomalies across all of these sets than GPT-2. This facilitates broader application of the paired perplexity approach than was previously possible, and suggests our approach is more sensitive to task-agnostic dementia-related linguistic anomalies than BERT/DistilBERT fine-tuning.

In contrast to impairing embeddings or individual attention layers, the maximum discriminating effect was achieved by impairing multiple attention layers (either combinatorially or cumulatively), which is consistent with prior observations that Transformer layers encode different syntactic and semantic linguistic features in multiple lower and middle layers \citep{jo-myaeng-2020-roles, jawahar-etal-2019-bert, lin-etal-2019-open}. Thus, impairing a single layer may not be enough to achieve the full effect.  Since both syntactic and semantic context is encoded in the Transformer decoder layers we expected to find different patterns of artificial impairment to be most effective in vastly different types of discourse represented by the DB and CCC datasets; however, we were surprised to find that only impairing the self-attention layers had the desired effect on the results in contrast to impairing embeddings or feed-forward network components.   

%Impairment of the lower layers had the greatest impact on ability to discriminate between language of patients with dementia and controls. Validation experiment results show our approach generalizes reasonably well across tasks and discourse types, where both cumulative and combinatorial methods achieve an average cross-set AUC and accuracy $\sim$ 0.7 on the ADReSS, DB and CCC datasets, while fine-tuned BERT leads to poor validity across discourse.

%However, our models do not perform as well on the AFPD dataset. Patients in this set had a mean MMSE score of 23.04, indicating relatively mild dementia. MMSE scores above 24 are commonly found in cognitively healthy participants, whereas scores between 21 and 24 indicate possible Mild Cognitive Impairment (MCI), and scores of 21 or less suggest moderate dementia \citep{Dick496,creavin2016mini}. Patients with dementia in the other three datasets are much more severely impaired (MMSE < 21) than those in the AFPD dataset, where differences between case and control language may be too subtle for accurate discrimination using  paired perplexity. This finding is also consistent with prior work showing that even manual analysis of linguistic, semantic, and discourse features of connected language samples cannot reliably discriminate between controls and patients with mild dementia, whereas these methods are more robust in discriminating between controls and patients with moderate/advanced dementia (see \citep{Mueller2018} for a comprehensive review).

The results presented in Table~\ref{tab:lan_stat_accumu} also align with previously published findings that both neural networks trained on language produced by participants with dementia, and the lexical-retrieval processes of patients affected by this condition are sensitive to lexical frequency effects  \citep{cohen-pakhomov-2020-tale,pekkala2013}. Our results suggest that impairing the self-attention mechanism in a Transformer artificial neural network may induce similar sensitivity to lexical frequency. By impairing the attention heads in a GPT-2, we observe significant differences in lexical frequency and TTR characteristics of the text generated by the GPT-2 and GPT-D, with the change in TTR ratio indicating that GPT-D has a greater tendency to produce repeated words when generating text, just as participants with dementia are more prone to repeat words in picture description tasks \citep{hier1985language}. 

% add a bit of row padding so the table does not look too crammed
%\renewcommand{\arraystretch}{1.2}

In other previous work  on the DB and the ADReSS datasets, the authors attempted to predict individual MMSE scores in addition to discriminating between cases and controls \citep{Yancheva2015UsingLF,bib:LuzHaiderEtAl20ADReSS}. We could not perform a comparable analysis in the current study on account of focusing on using the paired perplexity measure as a single threshold to distinguish between cases and controls, While predicting MMSE is not the main focus of our study, we did find negative correlations between the paired perplexity measures and the MMSE scores, providing additional evidence that artificially impairing the attention mechanism of the GPT-2 model simulates cognitive effects of dementia detectable in language.

%Our study is a computationally explicit evidence of a previous study
Our findings are also consistent with previous work indicating that Transformer models are able to predict neural responses during language comprehension and generalize well across various datasets and brain imaging modalities \citep{schrimpf2021neural}. Thus, our work is another step in the direction of achieving better understanding of the relationship between the inner workings of generative artificial neural language models and the cognitive processes underlying human language production. Impairing how contextual information is stored in the self-attention mechanism \textit{in silico} creates similar deficits to what is observed in dementia. The next important step is perhaps to investigate how contextual information encoding is impaired \textit{in vivo} in AD dementia.

The encouraging results on the CCC dataset point to the possibility of developing  a tool for analysing patients' daily spontaneous conversations in a task-agnostic fashion. Generalizable across tasks and domains and easy-to-interpret language-based instruments for detecting anomalies potentially consistent with dementia can be most useful in clinical situations where the patient or family member raise a concern about unexplained changes in cognition. A simple to administer (or self-administer) language-based instrument for objective confirmatory testing (either at a single point in time or over a period of time) would be helpful to a clinician working in an overburdened and time-constrained clinical environment (e.g., primary care) to be able to validate or refute those cognitive concerns with added confidence. It is critical, however, that the instrument used for confirmatory testing makes as few assumptions as possible regarding the person's linguistic background or communicative style, or the type of discourse used for analysis (i.e., picture description vs. conversation). 

The work presented here has several limitations. The sizes of the datasets are small compared to those typically encountered in open domain NLP tasks. In this paper, we did not focus on mild cognitive impairment but acknowledge that it is an important and active area of research that has shown promise in detecting early signs of dementia \citep{5710404,satt2014speech,calza2021linguistic}. Also, all datasets are in American English, which could limit the applicability of our models to dementia-related differences in other forms of English, and would certainly limit their applicability to other languages. In addition, behavioral characteristics including language anomalies can arise as a result of deficits in multiple brain mechanisms and, while they can contribute to a diagnosis of a neurodegenrative condition as a screening tool, they cannot be used in isolation to establish a definitive diagnosis. While GPT-D resembles language behaviors commonly observed in dementia patients, GPT-2 and GPT-D should not be considered as accurate and comprehensive representations of human language and cognition, or as models that capture features specific to various forms of neurodegeneration. Lastly, we also notice that the pre-trained LM is heavily gender-biased, a problem that we hope ongoing efforts to improve the fairness of AI (e.g. \citep{sheng-etal-2020-towards}) will address over time.

%,
\section{Conclusion}
We developed a novel approach to automated detection of linguistic anomalies in AD, involving deliberately degrading a pre-trained Transformer, with SOTA performance on the ADReSS test set, and generalization to language from conversational interviews. This, and the detection of dementia-related linguistic characteristics in text generated by GPT-D, suggests that our method is sensitive to task-agnostic linguistic anomalies in dementia, broadening  the scope of application of methods for automated detection of dementia beyond language from standardized cognitive tasks.

\section*{Acknowledgement}
This research was supported by grants from the National Institute on Aging (AG069792) and Administrative Supplement (LM011563-S1) from the National Library of Medicine  

\section*{Responsible NLP Research}
We followed the Responsible NLP Research checklist and ACL code of ethics for this work.

% Entries for the entire Anthology, followed by custom entries
\bibliography{acl-camera-ready}
\bibliographystyle{acl_natbib}

\section*{Appendix}
%\label{sec:appendix}

\begin{figure}[ht]
    \centering
    \includegraphics[width=0.4\textwidth]{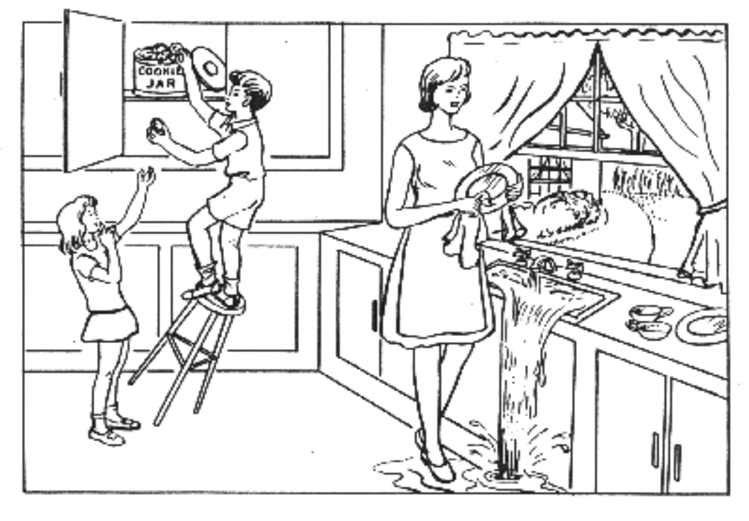}
    \caption{Cookie Theft picture stimulus.}
    \label{fig:cookie_theft}
\end{figure}

\begin{table}[h]
\resizebox{0.45\textwidth}{!}{\begin{tabular}{|p{3cm}|p{3.5cm}|p{3.5cm}|}
\hline
\textbf{Bird Transcript Sentence} & \textbf{GPT-2 Model Output} & \textbf{GPT-D Model Output} \\\hline
There are two children and their mother in the kitchen. & One of them is a little girl, but she's not very good at math or science so I don't really know what to do with her. & This is not a child, but they have no one parent's daughter or son of sister to father? brother family home with parents who can't let me know that I am sorry for my mommy baby daddy if you're just love them so much like it was good old time when we were talking about how many times \\\hline
The little boy has climbed up, on a three legged stool to get some cookies from the jar in the cupboard. & I'm so glad you're here, he says as we sit down at his kitchen table and watch him play with one of our favorite treats: chocolate chip cookie doughnuts! Oh my gosh! I say excitedly before turning back around for another bite that will make me feel good about what's been happening this whole time... & I don't know how do you like it's what is so good for kids love me and they're all i'm not sure no one day of year after years ago when people say 'I've been there was never ever again were always will be back then next time we got into my mom said she told her mother would tell them about their favorite thing that came out at nighttime before school started saying You can go home now or maybe even if your dad says he'll give him \\\hline
\end{tabular}}
\caption{Text generation examples by GPT-2 (small) compared to GPT-D created by the cumulative method of impairing the first 50\% attention heads of the first 9 layers.}
\label{tab:text_gene_comb}
\end{table}

\end{document}